\def\year{2020}\relax
\documentclass[letterpaper]{article} 
\usepackage{aaai20}  
\usepackage{times}  
\usepackage{helvet} 
\usepackage{courier}  
\usepackage[hyphens]{url}  
\usepackage{graphicx} 
\usepackage{graphicx}  
\title{3D Human Pose Estimation using  Spatio-Temporal Networks\\ with Explicit Occlusion Training}
\author{Yu Cheng,\textsuperscript{\rm 1}\thanks{Equal contribution} Bo Yang,\textsuperscript{\rm 2*} Bo Wang,\textsuperscript{\rm 2*} Robby T. Tan\textsuperscript{\rm 1,3}\\ 
\textsuperscript{\rm 1}National University of Singapore,  
\textsuperscript{\rm 2}Tencent Game AI Research Center,
\textsuperscript{\rm 3}Yale-NUS College \\
e0321276@u.nus.edu, \{brandonyang,bohawkwang\}@tencent.com, robby.tan@nus.edu.sg
}

\begin{document}

\maketitle

\begin{abstract}
Estimating 3D poses from a monocular video is still a challenging task, despite the significant progress that has been made in the recent years. Generally, the performance of existing methods drops when the target person is too small/large, or the motion is too fast/slow relative to the scale and speed of the training data. Moreover, to our knowledge, many of these methods are not designed or trained under severe occlusion explicitly, making their performance on handling occlusion compromised. Addressing these problems, we introduce a spatio-temporal network for robust 3D human pose estimation. As humans in videos may appear in different scales and have various motion speeds, we apply multi-scale spatial features for 2D joints or keypoints prediction in each individual frame, and multi-stride temporal convolutional networks (TCNs) to estimate 3D joints or keypoints. Furthermore, we design a spatio-temporal discriminator based on body structures as well as limb motions to assess  whether the predicted pose forms a valid pose and a valid movement. During training, we explicitly mask out some keypoints to simulate various occlusion cases, from minor to severe occlusion, so that our network can learn better and becomes robust to various degrees of occlusion. As there are limited 3D ground truth data, we further utilize 2D video data to inject a semi-supervised learning capability to our network. Experiments on public data sets validate the effectiveness of our method, and our ablation studies show the strengths of our network's individual submodules.\end{abstract}

\section{Introduction}
\label{sec:intro}
This paper focuses on 3D human pose estimation from a monocular RGB video. A 3D pose is defined as the 3D coordinates of pre-defined keypoints on humans, such as shoulder, pelvis, wrist, and etc. Recent top-down approaches \cite{hossain2018exploiting,Repnet,pavllo20183d,ChengICCV19} have shown promising results, where spatial features from individual frames are extracted to detect a target person and estimate the 2D poses, and temporal context is used to produce consistent 3D predictions. 

\begin{figure}[t]
\centering
\includegraphics[width=\linewidth]{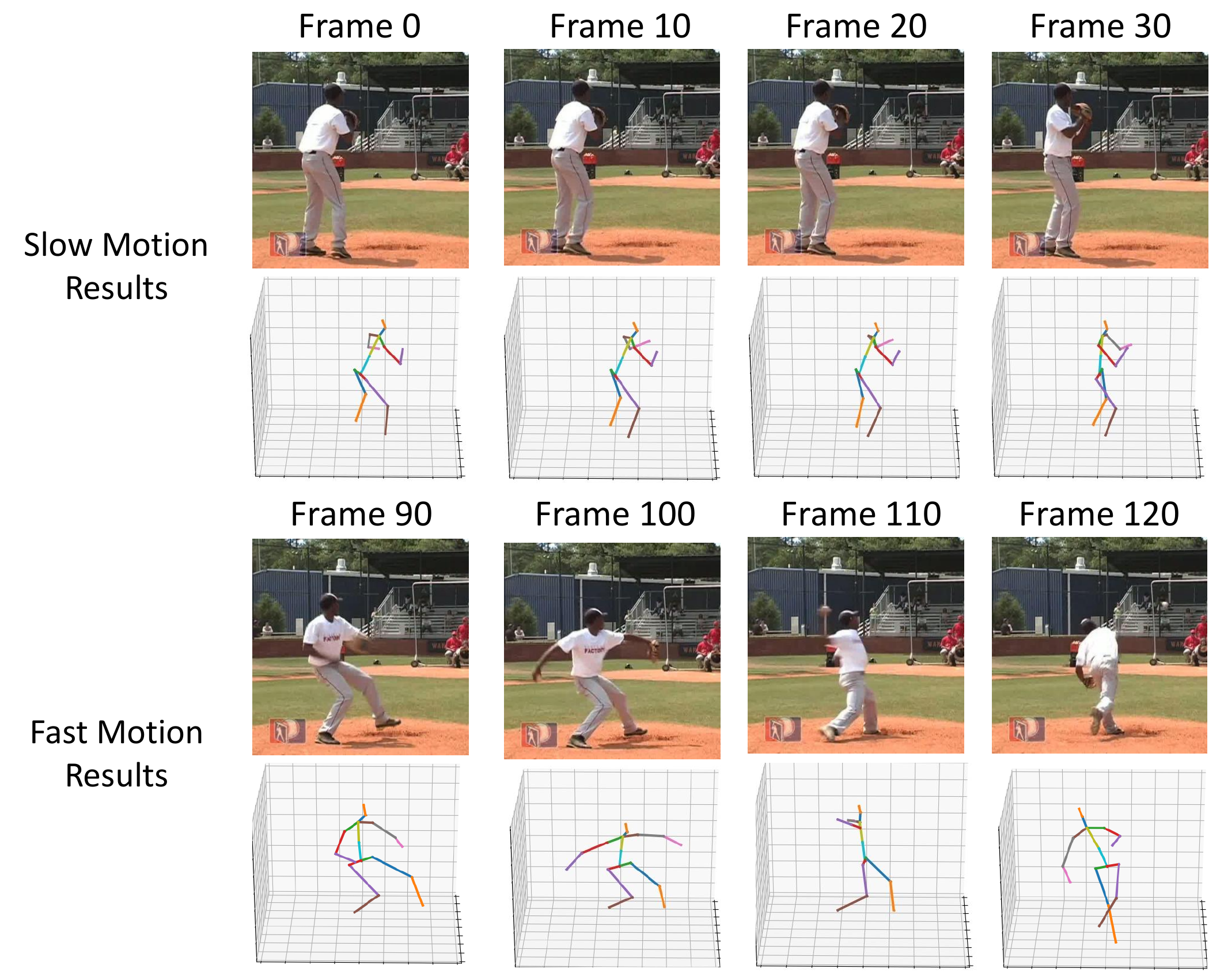}
\caption{Examples of our 3D human pose estimation under different movement speeds.}
\label{fig:multi-scale-example}
\end{figure}


However, we find that existing methods do not fully exploit  the spatial and temporal information available in videos. As a result, they suffer from the problem of large variations in  sizes and speeds of the target person in wild videos. In this paper, we address this problem.
First, we consider multi-scale features both spatially and temporally to deal with persons at various distances with different speeds of motions. We use the High Resolution Network (HRNet) \cite{sun2019deep} which exploits multi-scale spatial features to produce one heat map for each keypoint. Unlike most previous works \cite{newell2016stacked,pavllo20183d} that only use the peaks in the heat maps, we encode these maps into a latent space to incorporate more spatial information. Then, we apply temporal convolutional networks (TCNs) \cite{pavllo20183d} to these latent features with different strides, e.g., 1, 2, 4, and 8, and concatenate them together for prediction of the 3D poses. Figure \ref{fig:multi-scale-example} shows some expamples of our results. 

\begin{figure*}[t]
\centering
\includegraphics[width=\linewidth]{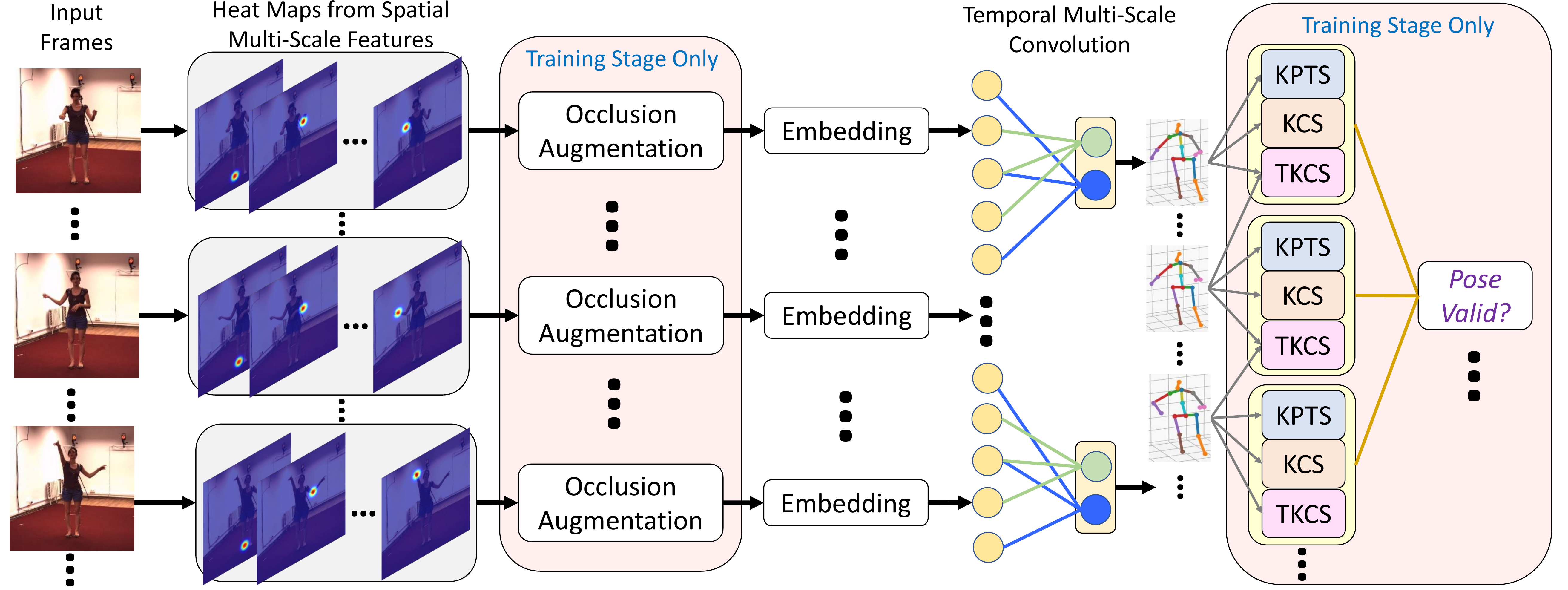}
\caption{Illustration for our framework. We only show two different temporal strides for clarity purpose. KPTS is short for keypoints; KCS is Kinematic Chain Space; TKCS means Temporal KCS.} 
\label{fig:framework}
\end{figure*}

Moreover, to reduce the risk of invalid 3D poses, we also utilize a discriminator in our framework like many previous works \cite{Yang20183DHP,Repnet,Chen_2019_CVPR,zhang2019phd}. Different from these methods' single frame based discriminators, we check the pose validity spatio-temporally. Our main reasoning is that valid poses in individual frames do not necessarily constitute a valid sequence. We extend the spatial KCS (Kinematic Chain Space) \cite{Repnet}, a successful single image descriptor for pose discriminator, and introduce a temporal KCS to represent motions of human joints. This temporal KCS descriptor is used by another TCN to check the validity of the estimated 3D pose sequence.

Finally, in order to deal with occlusion, during the training of our TCNs, we mask out some keypoints or frames by setting the corresponding heat maps to zero, as shown in Figure \ref{fig:framework}. There are two types of works that are similar to ours. One is the partial occlusion modeling by setting coordinates of some keypoints to zero~\cite{ChengICCV19}. The other is human dynamics, which only handles occlusion that happens in the end of a temporal window, since it predicts several future frames from given past frames' information~\cite{humanMotionKanazawa19,zhang2019phd}.
However, our approach can handle both partial and total occlusion cases in individual frames or in a sequence of frames. Hence, our method is more general in handling human 3D pose estimation under occlusion. Moreover, the occlusion module allows us to do semi-supervised learning that utilizes both 3D and 2D datasets.

As a summary, our contributions are as follows:
\begin{itemize}
    \item Incorporate multi-scale spatial and temporal features for robust pose estimation in video.
    \item Introduce a spatio-temporal discriminator to regularize the validity of a pose sequence.
    \item Perform diverse data augmentation for TCN to deal with different occlusion cases.
\end{itemize}

Experiments on public datasets show the efficacy of our contributions.

\section{Related Works}

Within the last few years, pose estimation has been undergoing rapid development with deep learning techniques~\cite{tompson2014joint,toshev2014deeppose,newell2016stacked,cao2018openpose,mehta2017vnect}. Researchers keep pushing the frontier of this field from different angles via better utilizing spatial or temporal information, learning human dynamics, pose regularization, and semi-supervised/self-supervised learning. 

To better utilize spatial information, some recent works focused on cross stage feature aggregation or multi-scale spatial feature fusion to maintain the high resolution in the feature maps~\cite{chen2018cascaded,sun2019deep,humanMotionKanazawa19}. Although this helps to improve the 2D estimators, there is an inherent ambiguity for inferring 3D human structure from a single 2D image. To overcome this limitation, some researchers further utilized temporal information in video~\cite{pavllo20183d,hossain2018exploiting,ChengICCV19,bertasius2019learning}, and showed obvious improvement. However, their fixed temporal scales limit their performance on videos with different motion speeds from the ones in training. 

To regularize predictions to be reasonable 3D human poses, pose discriminators have been proposed~\cite{Yang20183DHP,Repnet,Chen_2019_CVPR}. These methods utilize the idea of Generative Adversarial Networks (GAN) to check whether the estimated 3D pose is consistent with the pose distribution in the ground truth data. However, most of existing works focused on determining if a given 3D human pose is reasonable. Combining a series of reasonable 3D poses together does not make the whole series a reasonable human motion trajectory. As a result, we propose temporal KCS which checks both the spatial and temporal validity of 3D poses. 

To deal with partial occlusions, some techniques have been designed to recover occluded keypoints from unoccluded ones according to the spatial or temporal context \cite{radwan2013monocular,rogez2017lcr,de2018deep,guo2018occluded,ChengICCV19} or scene constraints~\cite{zanfir2018monocular,zanfir2018deep}. Some methods further introduced the concept of ``human dynamics''~\cite{humanMotionKanazawa19,zhang2019phd}, which predicts future human poses according to single or multiple existing frames in a video without any future frames. In real scenarios, we may have full, partial, or total occlusion for individual or continuous frames. Therefore, we introduce a method to integrate these two categories of methods into one unified framework by explicitly performing occlusion augmentation for all these cases during training.
Due to limited 3D human pose data, recent methods suggest to further utilize 2D human pose datasets in a semi-supervised or self-supervised fashion \cite{Repnet,wang2019selfsupervised,Kocabas_2019_CVPR,Chen_2019_CVPR}. They project estimated 3D pose back to 2D image space so that 2D ground-truth can be used for loss computation. Such approaches reduce the risk of over-fitting on small amount of 3D data. We also adopt this method and combine it with our explicit occlusion augmentation. 

\section{Methodology}
Our method belongs to the top-down pose estimation category. Given an input video, we first detect and track the persons by any state-of-the-art detector and tracker, such as Mask R-CNN \cite{he2017maskrcnn} and PoseFlow \cite{xiu2018poseflow}. Subsequently, we perform the pose estimation for each person individually.

\subsection{Multi-Scale Features for Pose Estimation}
Given a series of bounding boxes for a person in a video, we first normalize the image within each bounding box to a pre-defined fixed size, e.g., $256 \times 256$, and then apply High Resolution Networks (HRNet) \cite{sun2019deep} to each normalized image patch to produce $K$ heat maps, each of which indicates the possibility of certain human joint's location. The HRNet conducts repeated multi-scale
fusions by exchanging the information across the parallel multi-scale subnetworks. Thus, the estimated heat maps incorporate spatial multi-scale features to provide more accurate 2D pose estimations.

We concatenate the $K$ heat maps in each frame as a $K$-dimensional image $m_t$, where $t$ is the frame index, and apply an embedding network $f_E$ to produce a low dimensional representation as $r_t=f_E(m_t)$. Such embedding incorporates more spatial information from the whole heat maps than only using maps' peaks as most previous works do. The effectiveness of the embedding is shown in the ablation study in Table~\ref{tab:ablation}. 

\begin{figure}[t]
\centering
\includegraphics[width=\linewidth]{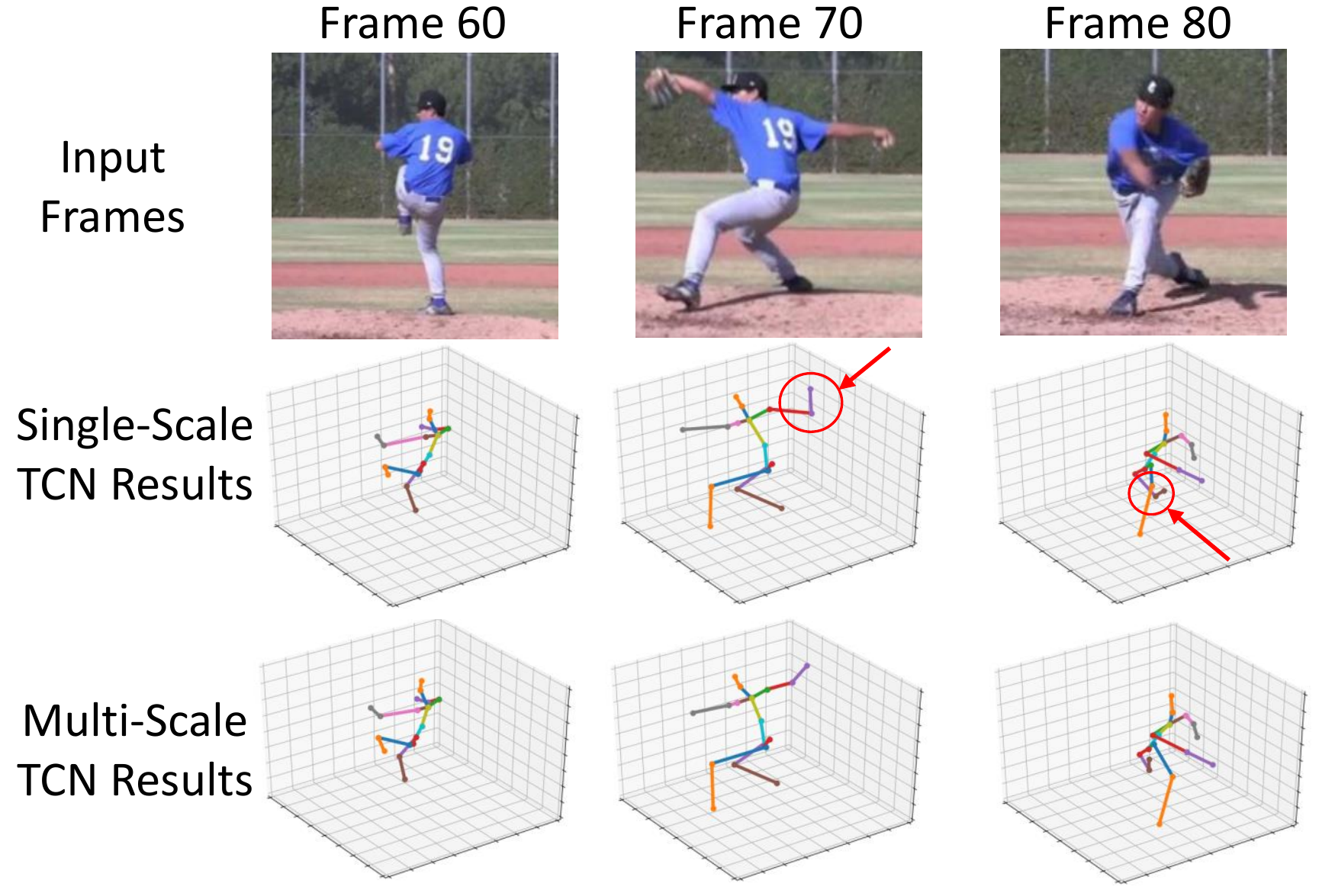}
\caption{Comparison of single-scale and multi-scale TCN results. Errors are labeled in red circles. The single-scale TCN fails to provide accurate predictions for fast motion frames.} 
\label{fig:multiscale}
\end{figure}

Given a sequence of heat map embeddings $\{r_t\}$, we apply TCN to them. As human motions may be fast or slow, we consider multi-scale features in the temporal domain. As shown in Figure \ref{fig:framework}, we apply TCN with temporal strides of 1, 2, 3, 5, 7 and concatenate these features for the final pose estimation. Such multi-scale features in both spatial and temporal domains enable our networks to deal with various scenarios. Figure~\ref{fig:multiscale} shows an example video clip with fast motion of playing baseball. We observe that multi-scale TCN is able to produce more accurate results than single-scale TCN.

We use both 3D dataset Human3.6M~\cite{h36m_pami} and 2D dataset Penn Action~\cite{penndataset} for training. Human3.6M has multi-view captured videos and 3D ground-truths, while PENN only has 2D ground-truths for visible keypoints. For Human3.6M data, the 3D MSE loss is defined as:
\begin{equation}
    L_{3d} = (X - X^{3D})^2,
\end{equation}
where $X$ is our predicted 3D coordinates for all keypoints, and $X^{3D}$ is the 3D ground truth. As Human3.6M data set provides videos from multiple views, we expect the 3D estimation results from different views should be the same after rotation alignment. So, we define the multi-view loss as:

\begin{equation}
    L_{mv} = (R_{v1 \rightarrow v2} X_{v1} - X_{v2} )^2,
\end{equation}
where $R_{v1 \rightarrow v2}$ is the rotation matrix from viewpoint 1 to viewpoint 2, and is precomputed from the ground-truth camera parameters. The $X_{v1}$ and $X_{v2}$ are the predicted 3D results in viewpoints 1 and 2.

For the 2D dataset, we project the 3D prediction to 2D space assuming orthogonal projection, and the 2D MSE loss is defined as: 
\begin{equation}
    L_{2d} = (Orth(X) - X^{2D})^2,
\label{eq:l2d}
\end{equation}
where $Orth(\cdot)$ is the orthogonal projection operator, and $X^{2D}$ is the 2D ground truth.

\subsection{Spatio-Temporal KCS Pose Discriminator}
\label{sec:discriminator}
To reduce the risk of generation of unreasonable 3D poses, we introduce a novel spatio-temporal discriminator to check the validity of a pose sequence, rather than just poses in individual frames like previous works~\cite{Yang20183DHP,Repnet,Chen_2019_CVPR}.

Among all single frame discriminators, the Kinematic Chain Space (KCS) used in~\cite{Repnet} is one of the most effective methods. Each bone, defined as the connection between two neighboring human keypoints such as elbow and wrist, is represented as a 3D vector $b_m$, indicating the direction from one keypoint to its neighbor. All such vectors form a $3 \times M$ matrix $B$, where $M$ is the predefined number of bones for a human structure. They use $\Psi = B^T B$ as the features for discriminator, where the diagonal elements in $\Psi$ indicate the square of bone length and other elements represent the weighted angle between two bones as an inner production.

\begin{figure}[t]
\centering
\includegraphics[width= \linewidth]{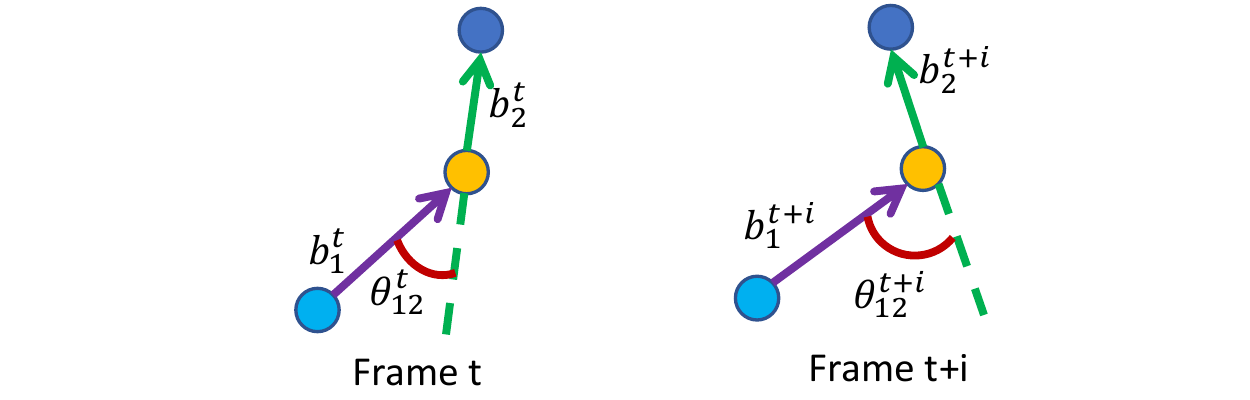}
\caption{Illustration for Temporal Kinematic Chain Space (TKCS) between two neighboring bones.} 
\label{fig:TKCS}
\end{figure}

Inspired by their spatial KCS, we introduce a Temporal KCS (TKCS) defined as:
\begin{equation}
    \Phi = B_{t+i}^TB_{t+i} - B_t^TB_t.
\end{equation}
where $i$ is the temporal interval between the KCS. The diagonal elements in $\Phi$ indicates the bone length changes, and other elements denote the change of angles between two bones. 
Figure \ref{fig:TKCS} shows an example of two neighboring bones $b_1$ and $b_2$. The spatial KCS measures the lengths of $b_1$ and $b_2$ as well as angles between them, $\theta_{12}$. The temporal KCS measures the bone length changes between two frames with temporal interval $i$, i.e., differences between $b_1^t$ and $b_1^{t+i}$ as well as $b_2^t$ and $b_2^{t+i}$, and the angle change between neighboring bones, i.e., difference between $\theta_{12}^t$ and $\theta_{12}^{t+i}$.

We concatenate the spatial KCS, temporal KCS, and the predicted keypoint coordinates, and then feed them to a TCN to build a discriminator. Such approach not only considers whether a pose is valid in individual frames, but also checks the validity of transitions across frames. We follow the procedure in the standard GAN to train the discriminator, and use it to produce a regularization loss for our predicted poses as $L_{gen}$.

In addition, to increase the robustness under different view angles, we introduce a rotational matrix as an augmentation to the generated 3D pose, as shown in the following equation:
\begin{equation}
    L'_{gen} = L_{gen}(RX),
\end{equation}
where $R$ is a rotational matrix $Rotation(\alpha, \beta, \gamma)$, and $\alpha$, $\beta$, $\gamma$ are rotational angles along $x$, $y$, and $z$ axis, respectively. As the rotational angles along $x$ and $z$ angles should be smaller compared with rotations along $y$ for normal human poses, in our experiments, $\beta$ is randomly sampled from $[-\pi, \pi]$ while $\alpha$ and $\gamma$ are sampled from $[-0.2\pi, 0.2\pi]$. 

The overall loss function for our training is defined as
\begin{equation} 
    L = L_{3d} + w_1 L_{mv} + w_2 L_{2d} + w_3 L'_{gen},
\end{equation}
where $w_1, w_2, w_3$ are set to $0.5$, $0.1$, $0.01$, respectively, and are fixed in all our experiments.

\begin{figure}[t]
\centering
\includegraphics[width=\linewidth]{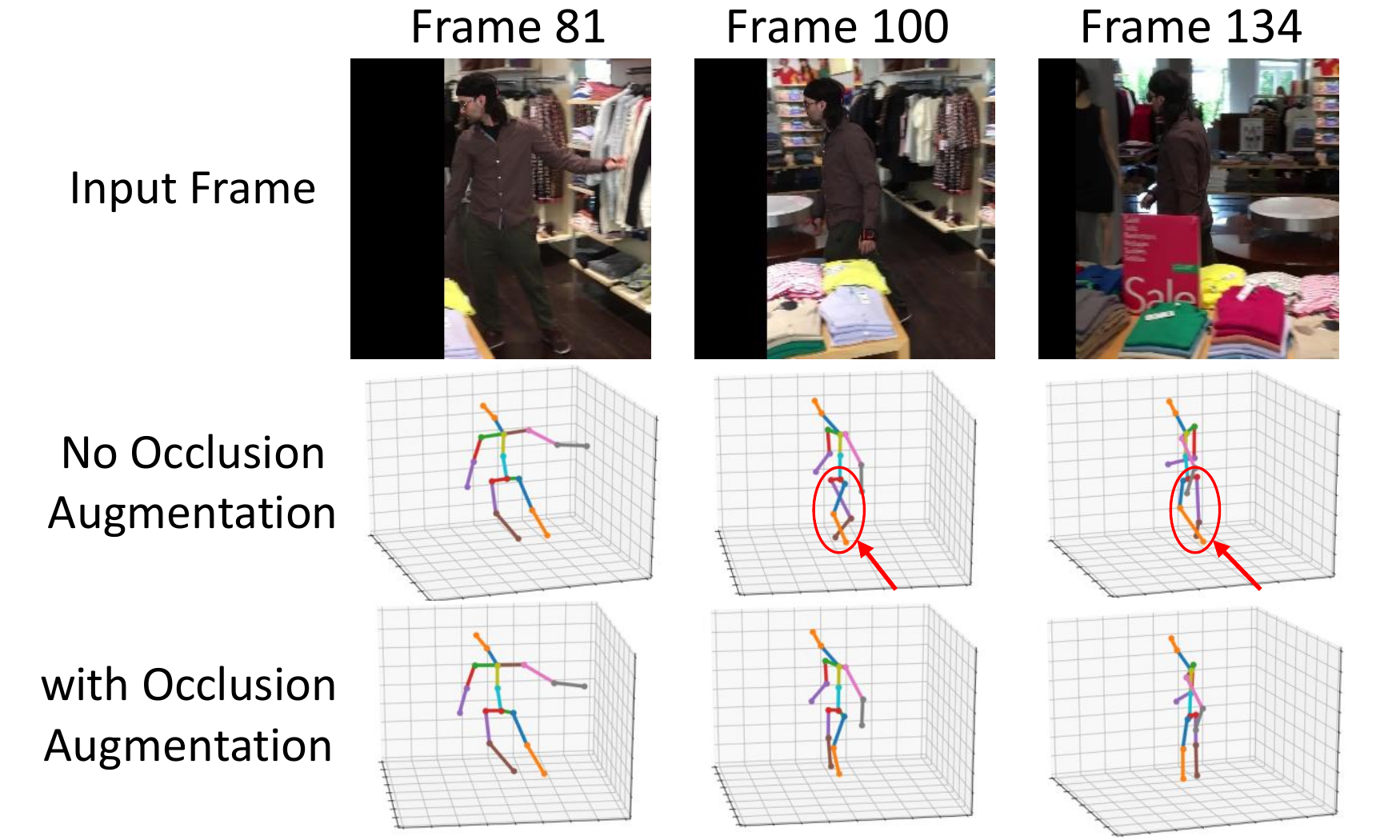}
\caption{Comparison of results from models trained with and without occlusion augmentation. Wrong estimations are labeled in red circles.}
\label{fig:occlusion}
\end{figure}

\subsection{Data Augmentation for Occlusions}
To make our approach capable of dealing with different occlusion cases, we perform data augmentation during the training. 

We use random masking of keypoints to simulate the occluded condition. Three types of occlusion are applied in the training process. The first type is the frame-wise occlusion. Given a sequence of heatmaps produced by the 2D keypoint estimator, we randomly mask several frames by setting their heatmaps to zero, indicating that the whole frame is occluded or has low confidence. Second, the point-wise occlusion is applied by randomly setting certain keypoints' heatmaps to zero. This simulates the scenario that certain keypoints are occluded. Third, we apply area occlusion by setting a virtual occluder area. The heatmaps of keypoints located within this area are set to zero. 

In addition, as the output of 2D pose estimator is not strictly Gaussian distribution, we introduce random noise to the heatmaps of the input sequence. To further improve the robustness under wrong detection cases, the points are randomly shifted or randomly swapped symmetrically. For example, the left knee is swapped with the right knee and the elbow point is sifted by 10 pixels. We expect the trained multi-scale TCN is able to recover the correct 3D pose using context information from partly wrong 2D estimations.

Note that, when the occlusion masks are all at the end of our TCN receptive field, it degrades to the human dynamics case, i.e., estimation of future poses without any future observation. Therefore, our framework is a more generalized approach for occlusion handling. We could predict human poses from temporal context information with or without meaningful observation in a few frames in the video clip. Figure \ref{fig:occlusion} demonstrates an example where occlusion augmentation helps to generate robust pose estimation results in a video clip where a target person is occluded. 

\section{Experiments}
\subsection{Experiment Settings}

\textbf{Data Sets.} Human3.6M \cite{h36m_pami} is a large 3D human pose dataset. It has 3.6 million images including eleven actors performing daily-life activities, and seven actors are annotated. The 3D ground-truth is provided by the mocap system, and the intrinsic/extrinsic camera parameters are known. Similar to some existing methods  \cite{hossain2018exploiting,pavllo20183d,pavlakos2018ordinal,Yang20183DHP}, we use subjects 1, 5, 6, 7, 8 for training, and the subjects 9 and 11 for evaluation. 

HumanEva-I is a relatively smaller dataset. Following the typical protocol \cite{martinez2017simple,hossain2018exploiting,pavllo20183d}, we use the same data division to train one model for all three actions (Walk, Jog, Box), and use the remaining data for testing. 
MPI-INF-3DHP~\cite{mono-3dhp2017} is a relatively new dataset that is captured in an indoor setting which is similar to the setting of Human3.6M. Following recent methods ~\cite{kanazawa2018end,pavlakos2018ordinal,Chen_2019_CVPR} that report their performance on this dataset, we utilize this dataset for quantitative evaluation.
3DPW~\cite{3DPW} is a new dataset contains multi-person outdoor scenes. We use the testing set of 3DPW to perform quantitative evaluation following~\cite{martinez2017simple,humanMotionKanazawa19}. 

\textbf{Evaluation protocols.}
We apply a few common evaluation protocols in our experiments. \textit{Protocol \#1} refers to the Mean Per Joint Position Error (MPJPE) which is the millimeters between the ground-truth and the predicted keypoints. \textit{Protocol \#2}, often called P-MPJPE, refers to the same error after applying alignment between the predicted keypoints and the ground-truth. Percentage of Correct 3D Keypoints (3D PCK) under $150mm$ radius is used for quantitative evaluation for MPI-INF-3DHP following~\cite{mono-3dhp2017}. To compare with other human dynamics/pose forecasting methods, mean angle error (MAE) is used following~\cite{jain2016structural}. 

\begin{table}
\footnotesize
\centering
  \begin{tabular}{c|c|c|c|c|c}
  \cline{1-6}
  \rule{0pt}{2.6ex}
  \textbf{Emb} & \textbf{T Len} & \textbf{T Strides}& \textbf{T Intvl} & P \#1 & P \#2 \\
    \cline{1-6}
    \rule{0pt}{2.6ex}
    64 & 64 & 1,2,3 & 1 & 58.3 & 44.2 \\
    128 & 64 & 1,2,3 & 1 & 46.7 & 36.1 \\
    256 & 64 & 1,2,3 & 1 & 43.1 & 33.8 \\
    512 & 64 & 1,2,3 & 1 & 42.6 & 33.4 \\
    1024& 64 & 1,2,3 & 1 & 42.9 & 33.6 \\
    \cline{1-6}
    512 & 8 & 1,2,3 & 1 & 50.2 & 40.1 \\
    512 & 16 & 1,2,3 & 1 & 46.9 & 36.0 \\
    512 & 32 & 1,2,3 & 1 & 44.0 & 33.9 \\
    512 & 64 & 1,2,3 & 1 & 42.6 & 33.4 \\
    512 & 128 & 1,2,3 & 1 & 42.9 & 33.7 \\
    \cline{1-6}
    512 & 64 & 1 & 1 & 45.4 & 35.9 \\
    512 & 64 & 1,2 & 1 & 44.3 & 34.8 \\
    512 & 64 & 1,2,3 & 1 & 42.6 & 33.4 \\
    512 & 64 & 1,2,3,5 & 1 & 41.8 & 32.1 \\
    512 & 64 & 1,2,3,5,7 & 1 & 41.2 & 31.5 \\
    \cline{1-6}
    512 & 64 & 1,2,3 & 1 & 42.6 & 33.4 \\
    512 & 64 & 1,2,3 & 3 & 43.1 & 33.7 \\
    512 & 64 & 1,2,3 & 5 & 44.0 & 34.4 \\
    \cline{1-6}
  \end{tabular}
  \caption{Parameter sensitivity test based on \textit{Protocol \#1 and \#2 of Human 3.6M dataset}. Emb stands for embedding dimension, T Len stands for Temporal length, T Strides stands for temporal strides, T Intvl stands for the temporal interval for TKCS.}
  \label{tab:sensitivity}
\end{table}

\subsection{Hyper-Parameter Sensitivity Analysis}
We conduct the sensitivity test of four hyper-parameters mentioned in this paper: embedding dimension for encoder, temporal length, temporal strides for TCN, and temporal interval for TKCS. The results are shown in Table~\ref{tab:sensitivity}. We find the best parameter settings by fixing three and adjusting the other one. To focus on understanding the influence of each parameter, semi-supervised learning using extra 2D data is disabled here. 

For the embedding dimension, we observe that within a reasonably large range, the performance is not affected significantly. The dimension $64$ is insufficient and results in large error. Within the range $256$ to $1024$, the errors only differ $0.4mm$, indicating that the model is insensitive to the setting of embedding dimension. 

For the temporal length, we test the range from $8$ to $128$. We can observe a steady reduction of errors until saturation at $128$. In addition, we adjust the temporal strides and find out that by adding more strides, the performance is improved and finally reaches $41.2mm$ with 5 strides compared to $45.4mm$ for single stride. We also test different temporal intervals for TKCS and observe interval $1$ produces the best performance. 

\begin{table}
\footnotesize
\centering
  \begin{tabular}{c|c|c}
  \cline{1-3}
  \rule{0pt}{2.6ex}
  \textbf{Method} & Protocol 1 & Protocol 2 \\
    \cline{1-3}
    \rule{0pt}{2.6ex}
    Base & 51.7 & 40.5 \\
    +embedding & 51.0 & 40.1 \\
    +multi-stride TCN & 48.6 & 37.6 \\
    +Multi-view loss & 47.3 & 36.9 \\
    +Spatial KCS & 44.9 & 34.0 \\
    +Temporal KCS & 41.2 & 31.5\\
    +2D Data & \textbf{40.1} & \textbf{30.7}\\
    \cline{1-3}
  \end{tabular}
  \caption{Ablation study on Human3.6M dataset under \textit{Protocol \#1 and \#2}. Best in bold.}
  \label{tab:ablation}
\end{table}

\begin{table*}[t]
\footnotesize
\centering
  \begin{tabular}{p{3.9cm}p{0.38cm}p{0.35cm}p{0.28cm}p{0.35cm}p{0.4cm}p{0.35cm}p{0.35cm}p{0.4cm}p{0.28cm}p{0.35cm}p{0.4cm}p{0.35cm}p{0.45cm}p{0.35cm}p{0.7cm}|p{0.4cm}}
  \cline{1-17}
    \rule{0pt}{2.6ex}
    \textbf{Method} & Direct & Disc. & Eat & Greet & Phone & Photo & Pose & Purch. & Sit & SitD. & Smoke & Wait & WalkD. & Walk & WalkT. & Avg\\
    \cline{1-17}
    \rule{0pt}{2.6ex}Fang et al. AAAI \shortcite{fang2018learning} & 50.1 & 54.3 & 57.0 & 57.1 & 66.6 & 73.3 & 53.4 & 55.7 & 72.8 & 88.6 & 60.3 & 57.7 & 62.7 & 47.5 & 50.6 & 60.4\\
    Yang et al. CVPR \shortcite{Yang20183DHP} & 51.5 & 58.9 & 50.4 & 57.0 & 62.1 & 65.4 & 49.8 & 52.7 & 69.2 & 85.2 & 57.4 & 58.4 & 43.6 & 60.1 & 47.7 & 58.6\\
    Hossain \& Little ECCV \shortcite{hossain2018exploiting} & 44.2 & 46.7 & 52.3 & 49.3 & 59.9 & 59.4 & 47.5 & 46.2 & 59.9 & 65.6 & 55.8 & 50.4 & 52.3 & 43.5 & 45.1 & 51.9\\
    Li et al. CVPR \shortcite{Li_2019_CVPR} & 43.8 & 48.6 & 49.1 & 49.8 & 57.6 & 61.5 & 45.9 & 48.3 & 62.0 & 73.4 & 54.8 & 50.6 & 56.0 & 43.4 & 45.5 & 52.7\\
    Chen et al. CVPR \shortcite{Chen_2019_CVPR} & - & - & - & - & - & - & - & - & - & - & - & - & - & - & - & 51.0 \\
    Wandt et al. CVPR \shortcite{Repnet} * & 50.0 & 53.5 & 44.7 & 51.6 & 49.0 & 58.7 & 48.8 & 51.3 & \underline{51.1} & 66.0 & 46.6 & 50.6 & \underline{42.5} & 38.8 & 60.4 & 50.9 \\
    Pavllo et al. CVPR \shortcite{pavllo20183d} & 45.2 & 46.7 & \underline{43.3} & 45.6 & 48.1 & 55.1 & 44.6 & 44.3 & 57.3 & 65.8 & 47.1 & \underline{44.0} & 49.0 & \textbf{32.8} & 33.9 & 46.8\\
    Cheng et al. ICCV \shortcite{ChengICCV19} & 
    \underline{38.3} & \underline{41.3} & 46.1 & \underline{40.1} & \underline{41.6} & \underline{51.9} & \underline{41.8} & \textbf{40.9} & 51.5 & \underline{58.4} & \underline{42.2} & 44.6 & \textbf{41.7} & 33.7 & \underline{30.1} & \underline{42.9}\\
    \cline{1-17}
    Our result & 
    \textbf{36.2} & \textbf{38.1} & \textbf{42.7} & \textbf{35.9} & \textbf{38.2} & \textbf{45.7} & \textbf{36.8} & \underline{42.0} & \textbf{45.9} & \textbf{51.3} & \textbf{41.8} & \textbf{41.5} & 43.8 & \underline{33.1} & \textbf{28.6} & \textbf{40.1} \\
    \cline{1-17}
  \end{tabular}
  \caption{Quantitative evaluation using MPJPE in millimeter between estimated pose and the ground-truth on Human3.6M under \textit{Protocol \#1}, no rigid alignment or transform applied in post-processing. Best in bold, second best underlined. * indicates ground-truth 2D labels are used.} 
  \label{tab:h3.6p1}
\end{table*}

\begin{table*}[t]
\footnotesize
\centering
  \begin{tabular}{p{3.9cm}p{0.38cm}p{0.35cm}p{0.28cm}p{0.35cm}p{0.4cm}p{0.35cm}p{0.35cm}p{0.4cm}p{0.28cm}p{0.35cm}p{0.4cm}p{0.35cm}p{0.45cm}p{0.35cm}p{0.7cm}|p{0.4cm}}
  \cline{1-17}
    \rule{0pt}{2.6ex}
    \textbf{Method} & Direct & Disc. & Eat & Greet & Phone & Photo & Pose & Purch. & Sit & SitD. & Smoke & Wait & WalkD. & Walk & WalkT. & Avg\\
    \cline{1-17}
    \rule{0pt}{2.6ex}Fang et al. AAAI \shortcite{fang2018learning} & 38.2 & 41.7 & 43.7 & 44.9 & 48.5 & 55.3 & 40.2 & 38.2 & 54.5 & 64.4 & 47.2 & 44.3 & 47.3 & 36.7 & 41.7 & 45.7\\
    Yang et al. CVPR \shortcite{Yang20183DHP} & \underline{26.9} & 30.9 & 36.3 & 39.9 & 43.9 & 47.4 & \textbf{28.8} & \underline{29.4} & \underline{36.9} & 58.4 & 41.5 & \textbf{30.5} & \textbf{29.5} & 42.5 & 32.2 & 37.7\\
    Hossain \& Little ECCV \shortcite{hossain2018exploiting} & 36.9 & 37.9 & 42.8 & 40.3 & 46.8 & 46.7 & 37.7 & 36.5 & 48.9 & 52.6 & 45.6 & 39.6 & 43.5 & 35.2 & 38.5 & 42.0\\
    Kocabas et al. CVPR \shortcite{Kocabas_2019_CVPR} & - & - & - & - & - & - & - & - & - & - & - & - & - & - & - & 45.0 \\
    Li et al. CVPR \shortcite{Li_2019_CVPR} & 35.5 & 39.8 & 41.3 & 42.3 & 46.0 & 48.9 & 36.9 & 37.3 & 51.0 & 60.6 & 44.9 & 40.2 & 44.1 & 33.1 & 36.9 & 42.6\\
    Wandt et al. CVPR \shortcite{Repnet} * & 33.6 & 38.8 & \underline{32.6} & 37.5 & 36.0 & 44.1 & 37.8 & 34.9 & 39.2 & 52.0 & 37.5 & 39.8 & 34.1 & 40.3 & 34.9 & 38.2 \\
    Pavllo et al. CVPR \shortcite{pavllo20183d} & 34.1 & 36.1 & 34.4 & 37.2 & 36.4 & 42.2 & 34.4 & 33.6 & 45.0 & 52.5 & 37.4 & 33.8 & 37.8 & \textbf{25.6} & \underline{27.3} & 36.5\\
    Cheng et al. ICCV \shortcite{ChengICCV19} & 28.7 & \underline{30.3} & 35.1 & \underline{31.6} & \underline{30.2} & \underline{36.8} & 31.5 & \textbf{29.3} & 41.3 & \underline{45.9} & \underline{33.1} & 34.0 & \underline{31.4} & 26.1 & 27.8 & \underline{32.8}\\
    \cline{1-17}
    Our result & \textbf{26.2} & \textbf{28.1} & \textbf{31.1} & \textbf{28.4} & \textbf{28.5} & \textbf{32.9} & \underline{29.7} & 31.0 & \textbf{34.6} & \textbf{40.2} & \textbf{32.4} & \underline{32.8} & 33.1 & \underline{26.0} & \textbf{26.1} & \textbf{30.7}\\
    \cline{1-17}
  \end{tabular}
  \caption{Quantitative evaluation using P-MPJPE in millimeter between estimated pose and the ground-truth on Human3.6M under \textit{Protocol \#2}. Procrustes alignment to the ground-truth is used in post-processing. Best in bold, second best underlined. * indicates ground-truth 2D labels are used.}
  \label{tab:h3.6p2}
\end{table*}

\subsection{Ablation Studies}
We conduct ablation studies to analyze each component of the proposed framework as shown in Table~\ref{tab:ablation}. 
As the baseline, we build a TCN to regress the 3D keypoints' positions based solely on their 2D coordinates $(x,y)$, which are obtained from the peaks in heatmaps from 2D pose detector. During TCN training, the 3D skeletons are also rotated along x, y, z axes as mentioned before. We use the standard MSE loss for the training.

We then add the modules one-by-one to perform ablation studies, including heat maps embedding, multi-stride TCN, multi-view loss, spatial KCS, temporal KCS, and 2D data semi-supervised learning. We see that by adding more modules, the performance steadily improves, validating the effectiveness of our proposed modules. The largest improvements come from multi-stride TCN, spatial KCS, and temporal KCS modules. Temporal multi-scale features increase the capability of the networks to deal with videos with different speeds of motions. Although the spatial KCS  constraints the pose validity at individual frames properly, our temporal KCS clearly further improves the performance, which demonstrates that checking the pose validity of a single frame itself is insufficient, and  checking the validity of the temporal pose sequence is necessary.

\begin{table*}
\footnotesize
\centering
  \begin{tabular}{p{2.9cm}|p{0.25cm}p{0.25cm}p{0.25cm}p{0.25cm}p{0.5cm}|p{0.25cm}p{0.25cm}p{0.25cm}p{0.25cm}p{0.5cm}|p{0.25cm}p{0.25cm}p{0.25cm}p{0.25cm}p{0.5cm}|p{0.25cm}p{0.25cm}p{0.25cm}p{0.25cm}p{0.5cm}}
  \cline{1-21}
  \rule{0pt}{2.6ex}
  \textbf{Actions} & \multicolumn{5}{c|}{Walking} & \multicolumn{5}{c|}{Eating} & \multicolumn{5}{c|}{Smoking} & \multicolumn{5}{c}{Discussion}\\
  \cline{1-21}
  \rule{0pt}{2.6ex}
  \textbf{Milliseconds} & 80 & 160 & 320 & 560 & 1000 & 80 & 160 & 320 & 560 & 1000 & 80 & 160 & 320 & 560 & 1000 & 80 & 160 & 320 & 560 & 1000 \\
  \cline{1-21}
  \rule{0pt}{2.6ex}Ghosh et al. \shortcite{ghosh2017learning} & 1.00 & 1.11 & 1.39 & 1.55 & 1.39 & 1.31 & 1.49 & 1.86 & 1.76 & 2.01 & 0.92 & 1.03 & 1.15 & 1.38 & 1.77 & 1.11 & 1.20 & 1.38 & 1.53 & \textbf{1.73} \\
  Martinez et al. \shortcite{martinez2017human} & 0.32 & 0.54 & 0.72 & 0.86 & 0.96 & \underline{0.25} & 0.42 & 0.64 & 0.94 & 1.30 & 0.33 & 0.60 & 1.01 & 1.23 & 1.83 & 0.34 & 0.74 & 1.04 & 1.43 & 1.75 \\
  Chiu et al. \shortcite{chiu2019action} & \textbf{0.25} & \textbf{0.41} & \textbf{0.58} & \textbf{0.74} & \textbf{0.77} & \textbf{0.20} & \textbf{0.33} & \textbf{0.53} & \textbf{0.84} & \underline{1.14} & \textbf{0.26} & \underline{0.48} & \textbf{0.88} & \textbf{0.98} & \underline{1.66} & \textbf{0.30} & \underline{0.66} & \underline{0.98} & \underline{1.39} & \underline{1.74} \\
  Our result & \underline{0.29} & \underline{0.48} & \underline{0.65} & \underline{0.79} & \underline{0.92} & \underline{0.25} & \underline{0.39} & \underline{0.58} & \underline{0.87} & \textbf{1.02} & 0.34 & \textbf{0.44} & \underline{0.90} & \underline{1.07} & \textbf{1.52} & 0.33 & \textbf{0.63} & \textbf{0.90} & \textbf{1.30} & 1.77 \\
  \cline{1-21}
  \end{tabular}
  \caption{Evaluation on Human3.6M dataset on human dynamics protocol. Mean angle error of predicted 3D poses after different time intervals is used following~\cite{martinez2017human,ghosh2017learning}. The milliseconds is the set future time for checking the performance. Best in bold, second best underlined.}
  \label{tab:dynamics}
\end{table*}

\begin{table}[h]
\footnotesize
\centering
  \begin{tabular}{p{3.0cm}|p{0.15cm}p{0.18cm}p{0.4cm}|p{0.15cm}p{0.15cm}p{0.4cm}|p{0.4cm}}
  \cline{1-8}
  \rule{0pt}{2.6ex}
  \textbf{Method} & \multicolumn{3}{c|}{Walking} & \multicolumn{3}{c|}{Jogging} & Avg \\
    \cline{1-8}
    \rule{0pt}{2.6ex}Pavlakos et al. \shortcite{pavlakos2018ordinal}\MakeUppercase{*} & 18.8 & 12.7 & 29.2 & 23.5 & 15.4 & 14.5 & 18.3\\
    Hossain et al. \shortcite{hossain2018exploiting} & 19.1 & 13.6 & 43.9 & 23.2 & 16.9 & 15.5 & 22.0\\
    Wang et al. \shortcite{wang2019selfsupervised} & 17.2 & 13.4 & \underline{20.5} & 27.9 & 19.5 & 20.9 & 19.9\\
    Pavllo et al. \shortcite{pavllo20183d} & 13.4 & \underline{10.2} & 27.2 & \underline{17.1} & 13.1 & 13.8 & 15.8\\
    Cheng et al. \shortcite{ChengICCV19} & \underline{11.7} & \textbf{10.1} & 22.8 & 18.7 & \textbf{11.4} & \textbf{11.0} & \underline{14.3}\\
    \cline{1-8}
    Our result & \textbf{10.6} & 11.8 & \textbf{19.3} & \textbf{15.8} & \underline{11.5} & \underline{12.2} & \textbf{13.5}\\
    \cline{1-8}
  \end{tabular}
  \caption{Evaluation on HumanEva-I dataset under \textit{Protocol \#2}. \textbf{Legend:} (*) uses extra depth annotations for ordinal supervision. Best in bold, second best underlined.}
  \label{tab:humanEvaI}
\end{table}

\begin{table}
\footnotesize
\centering
  \begin{tabular}{p{3.6cm}|p{0.4cm}}
  \cline{1-2}
  \rule{0pt}{2.6ex}
  \textbf{Method} & PCK \\
    \cline{1-2}
    \rule{0pt}{2.6ex}Mehta et al. 3DV \shortcite{mono-3dhp2017} & 72.5\\
    Yang et al. CVPR \shortcite{Yang20183DHP} &  69.0\\
    Chen et al. CVPR \shortcite{Chen_2019_CVPR} & 71.1\\
    Kocabas et al. CVPR \shortcite{Kocabas_2019_CVPR} & 77.5\\
    Wandt et al. CVPR \shortcite{Repnet} &  \underline{82.5}\\
    \cline{1-2}
    Our result & \textbf{84.1}\\
    \cline{1-2}
  \end{tabular}
  \caption{Evaluation on MPI-INF-3DHP dataset using 3D PCK. Best in bold, second best underlined. Only overlapped keypoints with Human3.6M are used for evaluation.}
  \label{tab:3dhp}
\end{table}

\begin{figure*}[h]
\centering
\makebox[\textwidth]{\includegraphics[width=\textwidth]{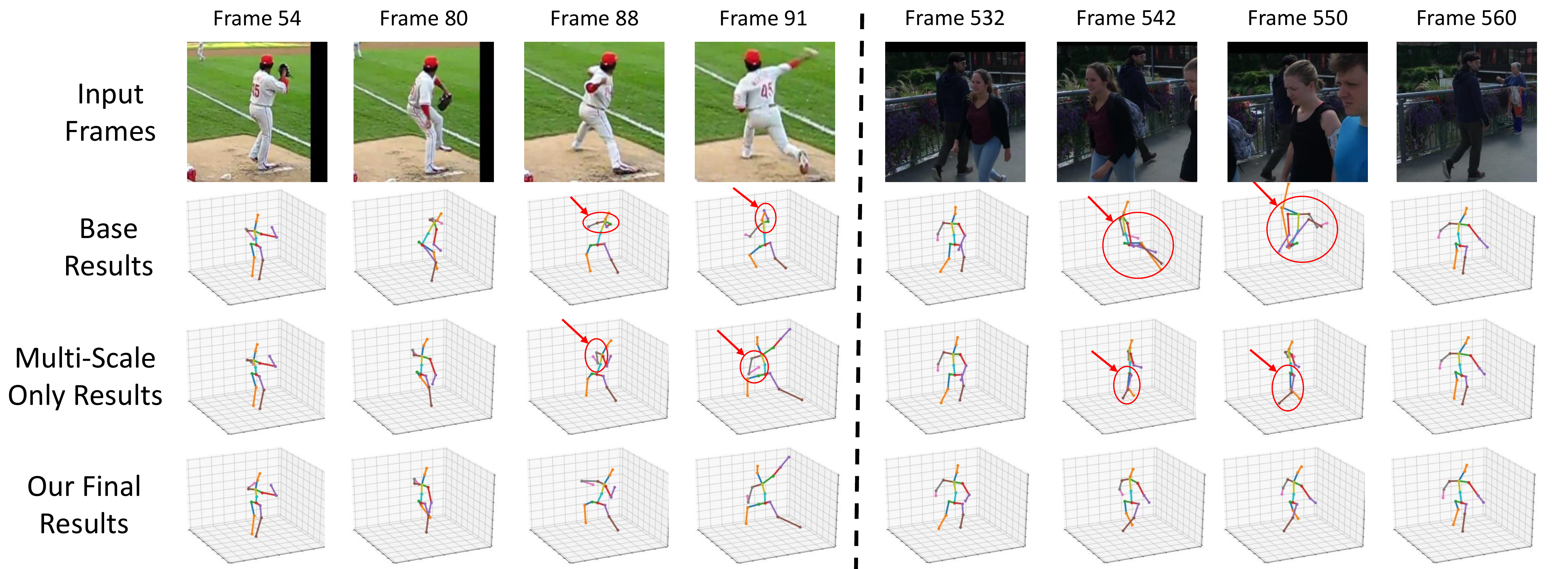}}
\caption{Examples of results from our whole framework compared with different baseline results. 
First row shows the images from two video clips;
second row shows the 3D results that uses baseline approach described in Ablation Studies; third row shows the 3D results that uses multi-scale temporal features without occlusion augmentation and spatio-temporal KCS; last row shows the results of the whole framework. Wrong estimations are labeled in red circles.}
\label{fig:qualitative_evaluation}
\end{figure*}

\subsection{Quantitative Results}
The experiment results on Human 3.6M are shown in Table~\ref{tab:h3.6p1} and Table~\ref{tab:h3.6p2} for \textit{Protocol \#1} and \textit{\#2}, respectively. The MPJPE is reduced by $2.8mm$ compared to previous work and yields an error reduction of $6.5\%$ . The P-MPJPE is reduced by $2.1mm$ and obtained $6.4\%$ error reduction. The performance on actions which already have low error rates is not improved significantly, but for those actions such as photo capturing and sitting down, the errors are reduced by $>5mm$. Since in these actions, occlusion happens frequently, more temporal information and effective pose regularization are needed for producing correct estimations. Considering existing methods almost get saturated on this dataset, our improvement is promising.

We also evaluate our model's potential on human dynamics which is targeted to predict several future frames' 3D skeleton. The performance is shown in Table~\ref{tab:dynamics}. Note that, \cite{chiu2019action} uses past 3D ground-truth keypoints as input for prediction, while our method does not use any ground-truth but takes images from video as input to estimate the keypoints first, and then predict the future 3D information, which is a more difficult task. Nevertheless, we still achieve similar performance compared with the state-of-the-art, which demonstrates the versatility of the proposed framework. Our method is not designed specifically for human dynamics prediction, but is a more generalized framework for pose estimation with or without observations (due to occlusion) in various scenarios.

Table~\ref{tab:humanEvaI} shows our evaluation results on HumanEva-I dataset and an improvement of $0.8mm$ is achieved in average, which implies an error reduction of $5.6\%$. For MPI-INF-3DHP dataset, we  use  only our model trained on Human3.6M dataset, but do not fine-tune or retrain on the 3DHP data set. Following existing methods~\cite{chiu2019action,martinez2017human}, we evaluate the Percentage of Correct 3D Keypoints (3D PCK) where points error under $150mm$ is considered correct (as the keypoints definitions are different in Human3.6M and 3DHP, we evaluate only the overlapped keypoints). As shown in Table~\ref{tab:3dhp}, even if we do not perform any re-training or fine-tuning, we still achieve an improvement of $1.6\%$ PCK, indicating the effectiveness of our approach. 

As above 3D human pose datasets contain mostly single-person indoor scenes, we also evaluate our framework on 3DPW, a new outdoor multi-person 3D poses dataset. Following~\cite{martinez2017simple,humanMotionKanazawa19} we do not train on 3DPW and only use its testing set for quantitative evaluation. The P-MPJPE value of our method on 3DPW testing set is 71.8, which outperforms the results, $157.0$ and $80.1$, reported in~\cite{martinez2017simple,humanMotionKanazawa19}.

\subsection{Qualitative Results}
Figure~\ref{fig:qualitative_evaluation} shows the 3D pose estimation results of the proposed framework compared with different baseline results. The first video clip (left four columns) shows a person playing baseball, which contains fast motion of limbs; the second video clip (right four columns) shows a person walking from left to right while some other people passing him, leading to occlusion. We use the same baseline method as used in the Ablation Studies. The baseline (the second row) fails on both video clips, because it cannot handle fast motion or occlusion. The results in the third row are from the method that uses multi-scale temporal features but without occlusion augmentation and spatio-temporal KCS based discriminator. We observe that it can handle the fast motion case to some extent but fails on the occlusion video, and the generated poses do not always satisfy anthropometrical constraints. The last row shows the results of our whole framework and it demonstrates our method can handle different motion speeds and various types of occlusion. 

\section{Conclusion}
In this paper, we present a new method  based on three major components: multi-scale temporal features, spatio-temporal KCS pose discriminator, and occlusion data augmentation. Our method can deal with videos with various motion speeds and different types of occlusion. The effectiveness of each component of our method is illustrated in the ablation studies. To compare with the state-of-the-art 3D pose estimation methods, we evaluate the proposed method on four public 3D human pose datasets with commonly used protocols and demonstrate our method's superior performance. Comparison with the human dynamics methods is provided as well to show our method is versatile and potentially can be used for other pose tasks, like pose forecasting.

{\small
\bibliography{AAAI-ChengY.4425.bib}
\bibliographystyle{aaai}
}

\end{document}